\documentclass[11pt]{article}

\usepackage[preprint]{acl}

\usepackage{times}
\usepackage{latexsym}  

\usepackage[T1]{fontenc}

\usepackage[utf8]{inputenc}

\usepackage{microtype}

\usepackage{inconsolata}

\usepackage{graphicx}

\usepackage{amsmath}
\usepackage{amssymb}
\usepackage{amsfonts}

\newcommand{\GR}{\mathcal{G}_{\mathrm{R}}}
\newcommand{\GM}{\mathcal{G}_{\mathrm{M}}}

\usepackage{bm}

\usepackage{booktabs}
\usepackage{colortbl}
\usepackage{xcolor}
\usepackage{multirow}
\usepackage{makecell}

\definecolor{darkgreen}{rgb}{0.0, 0.5, 0.0}
\definecolor{ourpurple}{RGB}{232, 228, 245}

\usepackage[most]{tcolorbox}
\usepackage{xcolor}
\usepackage{enumitem}

\definecolor{prompttitle}{RGB}{92,92,92}
\definecolor{promptborder}{RGB}{92,92,92}

\newtcolorbox{promptbox}[1]{
  enhanced,
  colback=white,
  colframe=promptborder,
  colbacktitle=prompttitle,
  coltitle=white,
  fonttitle=\bfseries\small,
  fontupper=\scriptsize,
  boxrule=1.0pt,
  arc=4pt,
  outer arc=4pt,
  left=7pt,
  right=7pt,
  top=5pt,
  bottom=5pt,
  title=#1,
  toptitle=3pt,
  bottomtitle=3pt,
  lefttitle=8pt,
  righttitle=8pt,
  before upper={
    \setlength{\parindent}{0pt}
    \setlength{\parskip}{2pt}
  }
}
\title{Reasoning or Memorization? Direction-Aware Diversity Exploration in LLM Reinforcement Learning}

\author{
Jiangnan Xia\textsuperscript{1},
Yucheng Shi\textsuperscript{2},
Yu Yang\textsuperscript{3},
Kishan Panaganti\textsuperscript{2},
Zhenwen Liang\textsuperscript{2},
Ninghao Liu\textsuperscript{4}
\\
\textsuperscript{1}University of Georgia \\
\textsuperscript{2}Tencent AI Lab \\
\textsuperscript{3}The Education University of Hong Kong \\
\textsuperscript{4}The Hong Kong Polytechnic University
}

\begin{document}
\maketitle
\begin{abstract}
Reinforcement learning has become a key paradigm for eliciting reasoning abilities in large language models, where exploration is crucial for discovering effective solution trajectories. 
Existing exploration methods typically encourage diversity in semantic or gradient spaces, without distinguishing what drives this diversity. 
A trajectory may appear novel because it follows a new reasoning process, or because it varies memorized patterns and shortcuts. 
Rewarding both cases equally may steer exploration toward memorization rather than genuine reasoning improvement. 
In this paper, we propose DiRL, a Direction-Aware Reinforcement Learning framework that anchors exploration to an internal reasoning-memorization direction of the policy.
Specifically, DiRL extracts this direction from model representations, constructs direction-weighted gradient features to characterize rollout updates, and shapes rewards to amplify reasoning-aligned exploration while suppressing memorization-aligned variations. 
DiRL integrates seamlessly into standard Group Relative Policy Optimization (GRPO). 
Extensive experiments on mathematical and general reasoning benchmarks demonstrate the
effectiveness of DiRL, showing significant improvements over various existing exploration methods. 
Our code is available at \url{https://anonymous.4open.science/r/DiRL-8F7C}

\end{abstract}

\section{Introduction}

Reinforcement learning (RL) has emerged as the primary paradigm for eliciting complex reasoning in large language models (LLMs) \cite{ladosz2022exploration,rafailov2023direct}. 
A central challenge in RL is exploration: the policy must sample diverse trajectories to avoid collapsing onto narrow, suboptimal modes \cite{shao2024deepseekmath, guo2025deepseek}. 
Recent methods encourage exploration by rewarding entropy or promoting diversity among sampled trajectories \cite{ouyang2022training, zhou2025evolving,liang2025can}. 
However, these approaches share a key limitation: \textbf{they treat all diversity as equally valuable, without distinguishing whether it arises from reasoning or memorization.}
A trajectory may appear novel because it explores a genuinely new reasoning path, or simply because it encounters patterns different from the model's memory. 
Existing exploration objectives reward both behaviors similarly.
As a result, exploration can amplify memorization-based variation alongside genuine reasoning, limiting its effectiveness for improving reasoning capability.

This limitation persists even as exploration metrics become more optimization-aware \cite{song2025outcome}. 
Earlier methods measure novelty in the semantic embedding space, encouraging trajectories that are distant from previous samples \cite{zhou2025evolving}. 
More recent work shifts this criterion to the gradient space, rewarding trajectories that induce distinct policy updates \cite{liang2025can}. 
While this shift better reflects how trajectories affect learning, a novel update can still reinforce memorization rather than improve reasoning. 
Thus, the key question is not merely whether a trajectory induces a novel update, but whether that update contributes to reasoning improvement.

\begin{figure*}[!t]
    \centering
    \includegraphics[scale=0.5]
     {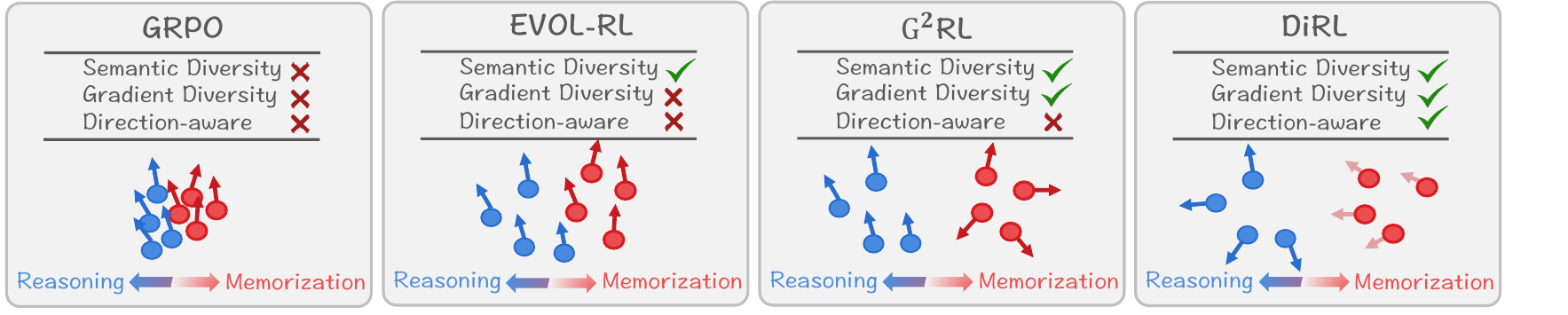}
    \caption{Comparison of different exploration strategies. Unlike existing diversity-based methods, DiRL selectively reinforces reasoning-aligned novelty while suppressing memorization-aligned variation.}
    \label{framework}
\end{figure*}

Recent studies suggest that reasoning and memorization correspond to distinct operational modes in LLMs \cite{nanda2023progress}. 
Models that perform well on familiar problems often fail under systematic perturbations, indicating reliance on memorized patterns rather than transferable reasoning \cite{dziri2023faith, berglund2024reversal}. 
Mechanistic analyses further show that these behaviors produce distinguishable signatures in the residual stream \cite{hernandez2024linearity, hong2025reasoning}. 
These findings offer a new opportunity for more effective RL exploration: instead of rewarding all diversity equally, exploration can \textbf{prioritize novelty associated with reasoning-aligned behaviors}.

In this paper, we propose DiRL, a Direction-Aware Reinforcement Learning framework that realizes this idea. 
DiRL aligns exploration with reasoning improvement by deriving a reasoning-memorization direction from the policy's own residual stream and using it as a geometric anchor for exploration. 
For each response, DiRL constructs a direction-weighted gradient feature, characterizing how the response would update the policy along this direction. 
The same direction further partitions responses into reasoning-aligned and memorization-aligned subgroups. 
An exploration score is then computed relative to the reasoning-aligned subgroup, ensuring that diversity is rewarded only when it expands reasoning rather than memorization. 
Finally, DiRL uses this score to shape the reward, amplifying reasoning-aligned responses while suppressing memorization-aligned ones. 
This design integrates seamlessly into GRPO \cite{shao2024deepseekmath}, converting generic diversity rewards into optimization signals that more directly support reasoning improvement.

We evaluate DiRL across math and general reasoning benchmarks, finding consistent improvements in pass@1, maj@16, and pass@16. 
Beyond raw accuracy, further analyses show that DiRL increases the proportion of reasoning-aligned rollouts, improves performance under symbolic perturbations, and introduces only modest computational overhead.
The main contributions are as follows: 
\begin{itemize}
    \item We introduce DiRL, a direction-aware RL framework which anchors exploration to the reasoning-memorization direction, 
    steering the policy toward reasoning rather than memorized patterns. 
   \item We develop a novel exploration mechanism that measures novelty in a direction-weighted gradient feature space and selectively amplifies reasoning-aligned exploration while suppressing memorization-aligned variations.
   \item Extensive experiments on multiple benchmarks demonstrate the effectiveness of our proposed framework DiRL. 
\end{itemize}

\section{Related Work}

\paragraph{Exploration in RL for LLM Reasoning.} 
Reinforcement learning with verifiable rewards has become a primary paradigm for training reasoning-capable LLMs \cite{ziegler2019fine, guo2025deepseek}, with group-relative algorithms like GRPO optimizing the policy from sparse binary signals \cite{shao2024deepseekmath}. The sparsity of this reward makes exploration crucial \cite{kearns2002near}. 
Early approaches encourage exploration by regularizing the output distribution \cite{mnih2016asynchronous}, such as adding a token-level entropy bonus to discourage early commitment to narrow modes \cite{ouyang2022training}. 
Subsequent work moves beyond token-level statistics and measures novelty at the trajectory level, with EVOL-RL \cite{zhou2025evolving} rewarding trajectories that are semantically distant in an external embedding space \cite{reimers2019sentence}. 
Recognizing that semantic distance may not align with optimization dynamics, G$^2$RL \cite{liang2025can} further shifts the diversity measurement into the policy's own gradient space, tying novelty directly to how each trajectory would update the model. 
Different from these methods that reward all diversity indiscriminately, 
DiRL anchors exploration to a reasoning-memorization direction, amplifying diversity that expands reasoning while suppressing diversity that drifts toward memorization.

\paragraph{Reasoning and Memorization in LLMs.} 
A growing body of work distinguishes genuine reasoning from memorization in LLM behavior and representations \cite{keysers2019measuring, li2025jointly}. 
Behavioral studies show that LLMs excelling on familiar problems often fail under systematic variations \cite{mirzadeh2025gsm}, sequence reversal \cite{berglund2024reversal}, or controlled input perturbations \cite{dziri2023faith}, suggesting reliance on memorized patterns rather than compositional reasoning. 
Beyond these behavioral evidences, mechanistic analyses identify multiple representational signatures of reasoning in LLMs. Mathematical reasoning is implemented through multi-step circuits in the residual stream \cite{hou2023towards, ye2025physics}, attention heads play distinct roles in knowledge retrieval and latent reasoning \cite{wang2022interpretability, elhage2021mathematical}, and reasoning- and memory-intensive inputs are linearly separable along a single direction in the residual stream \cite{hong2025reasoning}.
However, these findings have been used primarily as analytical tools for diagnosing model behavior or intervening at inference time. 
In contrast, DiRL incorporates such distinction into the training process, extracting the reasoning-memorization direction to guide reinforcement learning toward reasoning.

\section{Method}
We now introduce DiRL, a direction-aware reinforcement learning method for LLMs, which augments group-relative policy optimization (GRPO) with a reasoning-anchored exploration signal. 
First, we review GRPO and identify the role of exploration in the optimization process.
Second, we extract a reasoning-memorization direction from the policy's residual stream and uses this direction to shape exploration toward policy updates.
Then, we define a reasoning-anchored exploration score that measures novelty relative to reasoning-aligned responses.
Finally, we incorporate this score into GRPO through reward shaping to encourage reasoning driven exploration.

\subsection{Preliminaries: GRPO} 
\label{sec:GRPO}

Let $\pi_\theta$ be an autoregressive policy with parameters $\theta$, and $\pi_{\theta_\text{old}}$ denote the behavior policy used for rollout collection. For each prompt $x$ we sample a group $\mathcal{G}$ of $m$ candidate responses $\{y^{(i)}\}_{i=1}^m \sim \pi_{\theta_\text{old}}(\cdot \mid x)$, each scored by a verifier into a binary reward $r^{(i)} \in \{-1, +1\}$. GRPO \cite{shao2024deepseekmath} standardizes rewards within the group to obtain advantages,
\begin{equation}
A^{(i)} = \frac{r^{(i)} - \bar r}{\sigma_r + \varepsilon},
\end{equation}
where $\bar r = \tfrac{1}{m}\sum_{i=1}^m r^{(i)}$ and $\sigma_r^2 = \tfrac{1}{m}\sum_{i=1}^m (r^{(i)} - \bar r)^2$. 
The per-token importance ratio between the current and behavior policies is computed as,
\begin{equation}
\rho_t^{(i)}(\theta) = \frac{\pi_\theta(y_t^{(i)} \mid x, y_{<t}^{(i)})}{\pi_{\theta_\text{old}}(y_t^{(i)} \mid x, y_{<t}^{(i)})},
\end{equation}
and the GRPO objective combines a clipped surrogate with a KL penalty to a reference policy $\pi_\text{ref}$, 
\begin{equation}
\begin{aligned}
\mathcal{J}(\theta) = \mathbb{E}\Big[ &\tfrac{1}{m}\sum_i \tfrac{1}{L^{(i)}}\sum_t \min\!\big( \rho_t^{(i)} A^{(i)}, \\
&\quad \mathrm{clip}(\rho_t^{(i)}, 1-\epsilon, 1+\epsilon)\, A^{(i)} \big) \\
&- \beta\, D_\mathrm{KL}\!\big( \pi_\theta \parallel \pi_\text{ref} \big) \Big].
\end{aligned}
\end{equation}
The clipping range $\epsilon$ stabilizes policy updates, while the KL regularizer prevents drift from the reference policy. 
\textbf{DiRL changes the rewards used to construct the advantages}; the clipped surrogate and KL control remain unchanged.

\subsection{Reasoning-Memorization Direction Extraction}
\label{sec:axis}
DiRL first defines a \textit{reasoning-memorization direction} $\mathbf{k} \in \mathbb{R}^{d}$ in the policy residual stream. 
This direction serves as a reference for measuring whether a response aligns more strongly with reasoning or memorization behaviors.
Motivated by evidence that LLM activations differ between reasoning and factual retrieval~\cite{hong2025reasoning}, we construct two disjoint sets: $\mathcal{D}_+$ containing reasoning-intensive tasks (e.g., multi-step math problems), and $\mathcal{D}_-$ containing memory-intensive tasks (e.g., factual recall). 
We then compute $\mathbf{k}$ as the mean activation difference between the two sets:
\begin{equation}
\mathbf{k} = \frac{1}{|\mathcal{D}_+|}\!\sum_{x \in \mathcal{D}_+}\!
h^\star(x) \;-\; \frac{1}{|\mathcal{D}_-|}\!\sum_{x \in \mathcal{D}_-}\!
h^\star(x),
\label{eq:k}
\end{equation}
where $h^\star(x) \in \mathbb{R}^{d}$ is the final-layer hidden
state at the last prompt position. In autoregressive models, this representation aggregates information from the full input context. 

We compute $\mathbf{k}$ once using the initial policy parameters $\theta_0$ and keep it fixed throughout RL training. 
This design is practical because the KL regularization term in GRPO constrains policy drift, which preserves the overall geometry of the residual stream during optimization.

\subsection{Direction-Weighted Gradient Features}
\label{sec:phi}

For each response, we construct a gradient feature
$\Phi^{(i)} \in \mathbb{R}^{d}$ \textit{weighted by the reasoning direction}
$\mathbf{k}$, which summarizes how the response would update the policy along reasoning-relevant directions. 
Specifically, the policy's output layer maps each final-layer hidden state
$h_t \in \mathbb{R}^{d}$ to a distribution over the vocabulary via
$z_t = W^\top h_t$. This linearity gives rise to a \textit{per-token gradient
feature} \cite{liang2025can}:
\begin{equation}
\phi_t = W\big(e(y_t) - p_t\big) \in \mathbb{R}^{d},
\end{equation}
where $W \in \mathbb{R}^{V \times d}$ is the output projection
matrix, $e(y_t)$ is the one-hot encoding of the sampled token, and $p_t$
denotes the predicted distribution at position $t$. 
Each $\phi_t$ describes how position $t$ would update the policy, and can be computed directly from the forward pass without backpropagation. 
Moreover, the full parameter gradient at each layer factors through $\phi_t$
(Appendix~\ref{app:factorization}), so angular comparisons among
aggregated $\phi_t$ vectors provide a principled approximation of how different responses influence policy updates. 

However, a single response may contain both reasoning and memorization tokens. Uniformly summing $\phi_t$ across positions therefore mixes reasoning-related and memorization-related update signals. 
To emphasize reasoning-relevant updates, we weight each token
according to how its hidden state aligns with $\mathbf{k}$:
\begin{equation}
\alpha_t = \sigma\!\big(\langle h_t,\, \mathbf{k} \rangle\big),
\label{weighted-alpha}
\end{equation}
where $\sigma$ is the sigmoid function. Tokens whose hidden states align more strongly with the reasoning side of $\mathbf{k}$ receive larger weights, while tokens aligned with the memorization side receive smaller weights. 
The final direction-weighted gradient feature $\Phi^{(i)}$ is computed as:
\begin{equation}
\Phi^{(i)} = \sum_{t=1}^{L^{(i)}} \alpha_t\, \phi_t.
\label{weighted-eq}
\end{equation}
With this weighting, angular differences among $\Phi$ vectors primarily reflect diversity in reasoning.
These features form the basis of the exploration score in the next subsection.

\subsection{Reasoning-Anchored Exploration Score}
\label{sec:novelty}

With the gradient features $\Phi^{(i)}$, we assign each response an exploration score that measures how much novel information it contributes. 
The key question is how novelty should be defined, namely, which responses should serve as the reference set.
This choice of \emph{reference set} is important, because measuring novelty against the entire rollout group would reward any response that differs from its peers, regardless of whether the difference comes from reasoning or memorization (only the former contributes to reasoning-oriented exploration). 

DiRL addresses this by dividing responses into two subgroups, and anchoring novelty computation to the reasoning subgroup. 
Specifically, we project each response's mean hidden state onto $\mathbf{k}$, 
\begin{equation}
s^{(i)} = \Big\langle \frac{1}{L^{(i)}}\sum_t h_t^{(i)},\; \mathbf{k}
\Big\rangle .
\label{eq-subgroup}
\end{equation}
This partitions the rollout group $\mathcal{G}$ into a \textbf{reasoning-aligned subgroup} $\GR = \{i : s^{(i)} > 0\}$ and a \textbf{memorization-aligned subgroup} $\GM = \{i : s^{(i)} \le 0\}$. 

The exploration score is built from two quantities. 
The first is the \emph{cosine similarity} between two responses' gradient features, which captures directional differences in how they would update the policy, while being invariant to gradient magnitudes, 
\begin{equation}
S_{i,j} = \Big\langle \frac{\Phi^{(i)}}{\|\Phi^{(i)}\|},\;
\frac{\Phi^{(j)}}{\|\Phi^{(j)}\|} \Big\rangle.
\end{equation}
The second is a \emph{reward-weighted coefficient} that assigns larger weights to correct reference responses than incorrect ones,
\begin{equation}
w_{i,j} = \frac{\exp(r^{(j)})\,\mathbb{1}\{j \neq i\}}
{\sum_{k \neq i} \exp(r^{(k)}) + \varepsilon}.
\end{equation}
Based on $S_{i,j}$ and $w_{i,j}$, we compute the exploration score as,
\begin{equation}
\nu^{(i)} = \sqrt{\max\!\Big(1 - \sum_{j \in \mathcal{R}_i} w_{i,j}\,
S_{i,j}^2,\; 0\Big)}, \quad
\label{eq-score}
\end{equation}
where $\mathcal{R}_i$ represents the reference set. 

We want to encourage diversity \emph{within} the reasoning subgroup: each reasoning response should bring an update direction the others have not yet covered. 
For a reasoning response $i \in \mathcal{G}_R$, $\mathcal{R}_i=\mathcal{G}_R \setminus \{i\}$ is its reasoning peers (excluding itself). 
A high $\nu^{(i)}$ means this response induces a policy update direction not covered by other reasoning-aligned responses. 
On the other hand, for a memorization-aligned response $i \in \mathcal{G}_M$, we set $\mathcal{R}_i=\mathcal{G}_R$. Here, a high $\nu^{(i)}$ means the response lies far from reasoning-aligned update directions, so its training signals should be suppressed. 
If $\mathcal{R}_i$ is empty, we set $\nu^{(i)}=0$. 
Because the score $\nu^{(i)}$ may have different ranges across $\mathcal{G}_R$ and $\mathcal{G}_M$, we apply min-max normalization separately within each subgroup to obtain $\bar{\nu}^{(i)} \in [0, 1]$, yielding a comparable exploration signal across the two subgroups.

\subsection{Direction-Aware Reward Shaping}
\label{sec:shaping}

The exploration score $\nu^{(i)}$ tells us, for each response, whether its update direction strengthens or weakens reasoning. 
To incorporate this signal into policy optimization, we use it to shape the original reward $r^{(i)}$ into $\tilde{r}^{(i)}$:
\begin{equation}
\tilde{r}^{(i)} =
\begin{cases}
r^{(i)} + \lambda_{+}\, \bar{\nu}^{(i)} & \text{if } i \in \mathcal{G}_R, \\
r^{(i)} - \lambda_{-}\, \bar{\nu}^{(i)} & \text{if } i \in \mathcal{G}_M,
\end{cases}
\label{eq12}
\end{equation}
where $\lambda_{+}, \lambda_{-} \geq 0$ controls the shaping strength. 
The opposite signs reflect a single principle: \emph{amplify update directions that move the policy toward reasoning, reduce those that move it
away}. 

Combining $r^{(i)} \in \{-1, +1\}$ with subgroup membership yields
four trajectory types:
\begin{itemize}
\item \emph{Reasoning + correct} ($i\in\mathcal{G}_R, r^{(i)} = +1$): $\tilde{r}^{(i)}\geq 1$. Correct responses with novel reasoning directions receive larger rewards. 
\item \emph{Reasoning + wrong} ($i \in \mathcal{G}_R, r^{(i)} = -1$): 
$\tilde{r}^{(i)} \geq -1$. The penalty is softened, since a wrong but novel reasoning attempt still contributes useful update direction. 
\item \emph{Memorization + correct} ($i \in \mathcal{G}_M, r^{(i)} =
+1$): 
$\tilde{r}^{(i)} < 1$. Correct responses that rely on memorization updates instead of reasoning receive smaller rewards, discouraging memorized shortcuts.
\item \emph{Memorization + wrong} ($i \in \mathcal{G}_M, r^{(i)} =
-1$): $\tilde{r}^{(i)} \leq -1$. A wrong memorization response far from reasoning (hallucination signal) is penalized more strongly.
\end{itemize}

To keep the reward scale bounded, we clip $\tilde{r}^{(i)}$ to
$[-c, c]$ before feeding it into the GRPO advantage standardization. 
The resulting shaped rewards increase the advantages of reasoning-aligned responses and reduce those of memorization-aligned responses, gradually steering policy optimization toward reasoning-oriented exploration.

\section{Experiments}
\subsection{Experimental Setup}

\paragraph{LLM Backbones and Training.} 
We conduct experiments on Qwen3-1.7B-Base and Qwen3-4B-Base, training each model on the 7.5k problems of the MATH dataset~\cite{hendrycks2021measuring} with a rule-based verifier providing binary rewards $r^{(i)} \in \{-1, +1\}$. The full training hyperparameters and infrastructure details are provided in Appendix~\ref{app:hyperparams}.

\paragraph{Constructing the Reasoning Direction.}
DiRL extracts the reasoning direction $\mathbf{k}$ once before RL training via Eq.(\ref{eq:k}), using two prompt sets that contrast reasoning- and memory-intensive behaviors. 
To prevent $\mathbf{k}$ from collapsing to features specific to any single domain or language, we draw $\mathcal{D}_+$ from a diverse mix of reasoning-intensive sources: 
GSM8K \cite{cobbe2021training}, MGSM \cite{shi2022language}, the reasoning portion of MMLU-Pro \cite{wang2024mmlu}, and a reasoning-labeled split of MATH (MATH-R); 
$\mathcal{D}_-$ combines their memory-intensive counterparts: PopQA \cite{mallen2023not}, the humanities portion of C-Eval \cite{huang2023c}, the memory portion of MMLU-Pro, and MATH-M. 
The MATH splits are obtained by prompting GPT-4o to assign each problem a reasoning score and partitioning at the threshold $0.5$, these labels are used solely for extracting $\mathbf{k}$ and never enter the RL reward. Details are provided in Appendix~\ref{app:math_split}. 

\paragraph{Evaluation.}
Our main results are reported on four mathematical reasoning benchmarks of increasing difficulty: MATH500, AMC, AIME24, and AIME25. 
For each prompt, we report pass@1, maj@16, and pass@16. pass@$k$ is the fraction of prompts for which at least one of $k$ samples is verified correct; maj@16 is the majority-vote accuracy 
over 16 samples, where ties are broken uniformly. All numbers are reported as percentages.


\paragraph{Baselines.}
We compare DiRL against four exploration strategies: GRPO \cite{shao2024deepseekmath} uses no exploration bonus; Entropy Bonus \cite{mnih2016asynchronous} adds an entropy regularizer to the policy loss; EVOL-RL \cite{zhou2025evolving} uses external sentence embeddings to encourage semantic diversity; and G$^2$RL \cite{liang2025can} rewards novelty in the policy's gradient feature space. 



\begin{table*}[t]
\caption{Main results on MATH500, AMC, AIME24, and AIME25 with Qwen3-1.7B-Base and Qwen3-4B-Base.}
\centering
\small
\setlength{\tabcolsep}{4pt}
\renewcommand{\arraystretch}{1.0}
\definecolor{ourpurple}{RGB}{232, 228, 245}
\definecolor{gaingreen}{RGB}{0, 130, 0}

\newcommand{\ph}{\phantom{$_{+0.0}$}}

\resizebox{\textwidth}{!}{%
\begin{tabular}{l@{\hskip 6pt} ccc @{\hskip 6pt} ccc @{\hskip 6pt} ccc @{\hskip 6pt} ccc}
\toprule
& \multicolumn{3}{c}{\textbf{MATH500}} & \multicolumn{3}{c}{\textbf{AMC}} & \multicolumn{3}{c}{\textbf{AIME24}} & \multicolumn{3}{c}{\textbf{AIME25}} \\
\cmidrule(lr){2-4} \cmidrule(lr){5-7} \cmidrule(lr){8-10} \cmidrule(lr){11-13}
\textbf{Model} & pass@1 & maj@16 & pass@16 & pass@1 & maj@16 & pass@16 & pass@1 & maj@16 & pass@16 & pass@1 & maj@16 & pass@16 \\
\midrule
\multicolumn{13}{l}{\textit{Qwen3-1.7B-Base}} \\
GRPO          & 64.6\ph & 73.7\ph & 89.1\ph & 37.6\ph & 53.0\ph & 77.2\ph & 6.5\ph  & 12.9\ph & 25.3\ph & 4.4\ph  & 7.8\ph  & 21.6\ph \\
Entropy Bonus & 65.3\ph & 74.5\ph & 89.9\ph & 38.4\ph & 52.9\ph & 74.4\ph & 7.6\ph  & 12.6\ph & 26.6\ph & 4.2\ph  & 7.4\ph  & 22.7\ph \\
EVOL-RL       & 65.3\ph & 75.2\ph & 89.3\ph & 38.2\ph & 53.1\ph & 79.4\ph & 6.9\ph  & 13.5\ph & 24.5\ph & 5.4\ph  & 6.7\ph  & 19.7\ph \\
G$^2$RL       & 67.0\ph & 78.6\ph & 91.6\ph & 40.5\ph & 53.8\ph & 82.5\ph & 9.7\ph  & 17.4\ph & 27.4\ph & 6.0\ph  & 9.9\ph  & 23.4\ph \\
\rowcolor{ourpurple}
\textbf{DiRL (Ours)}
& \textbf{68.5}$_{\textcolor{gaingreen}{+3.9}}$
& \textbf{79.4}$_{\textcolor{gaingreen}{+5.7}}$
& \textbf{92.1}$_{\textcolor{gaingreen}{+3.0}}$
& \textbf{43.6}$_{\textcolor{gaingreen}{+6.0}}$
& \textbf{55.5}$_{\textcolor{gaingreen}{+2.5}}$
& \textbf{83.4}$_{\textcolor{gaingreen}{+6.2}}$
& \textbf{10.9}$_{\textcolor{gaingreen}{+4.4}}$
& \textbf{18.2}$_{\textcolor{gaingreen}{+5.3}}$
& \textbf{29.7}$_{\textcolor{gaingreen}{+4.4}}$
& \textbf{7.3}$_{\textcolor{gaingreen}{+2.9}}$
& \textbf{11.7}$_{\textcolor{gaingreen}{+3.9}}$
& \textbf{28.3}$_{\textcolor{gaingreen}{+6.7}}$ \\
\midrule
\multicolumn{13}{l}{\textit{Qwen3-4B-Base}} \\
GRPO          & 77.4\ph & 82.2\ph & 93.1\ph & 53.0\ph & 66.3\ph & 86.4\ph & 12.3\ph & 19.8\ph & 32.5\ph & 10.4\ph & 18.8\ph & 32.8\ph \\
Entropy Bonus & 78.3\ph & 84.8\ph & 93.7\ph & 53.7\ph & 67.8\ph & 88.8\ph & 13.8\ph & 22.2\ph & 36.3\ph & 12.1\ph & 23.5\ph & 37.6\ph \\
EVOL-RL       & 79.5\ph & 85.3\ph & 94.8\ph & 53.1\ph & 67.1\ph & 87.8\ph & 14.4\ph & 20.9\ph & 35.7\ph & 12.3\ph & 24.0\ph & 35.1\ph \\
G$^2$RL       & 79.3\ph & 86.5\ph & 94.7\ph & 54.1\ph & 68.4\ph & 90.4\ph & 17.0\ph & 25.4\ph & 40.5\ph & 15.6\ph & 25.7\ph & 42.0\ph \\
\rowcolor{ourpurple}
\textbf{DiRL (Ours)}
& \textbf{83.0}$_{\textcolor{gaingreen}{+5.6}}$
& \textbf{87.7}$_{\textcolor{gaingreen}{+5.5}}$
& \textbf{95.9}$_{\textcolor{gaingreen}{+2.8}}$
& \textbf{61.0}$_{\textcolor{gaingreen}{+8.0}}$
& \textbf{75.6}$_{\textcolor{gaingreen}{+9.3}}$
& \textbf{94.5}$_{\textcolor{gaingreen}{+8.1}}$
& \textbf{19.6}$_{\textcolor{gaingreen}{+7.3}}$
& \textbf{26.4}$_{\textcolor{gaingreen}{+6.6}}$
& \textbf{44.0}$_{\textcolor{gaingreen}{+11.5}}$
& \textbf{18.3}$_{\textcolor{gaingreen}{+7.9}}$
& \textbf{27.0}$_{\textcolor{gaingreen}{+8.2}}$
& \textbf{45.8}$_{\textcolor{gaingreen}{+13.0}}$ \\
\bottomrule
\end{tabular}%
}

\label{tab:main_results}
\end{table*}


\begin{table}[t]
\centering
\small
\setlength{\tabcolsep}{5pt}
\renewcommand{\arraystretch}{1.0}
\caption{Generalization results on GPQA and MMLU-Pro, reported in pass@1 (\%).}
\label{tab:generalization}
\resizebox{\columnwidth}{!}{%
\begin{tabular}{lcccc}
\toprule
& \multicolumn{2}{c}{GPQA} & \multicolumn{2}{c}{MMLU-Pro} \\
\cmidrule(lr){2-3} \cmidrule(lr){4-5}
Method &
{\scriptsize Qwen3-1.7B} &
{\scriptsize Qwen3-4B} &
{\scriptsize Qwen3-1.7B} &
{\scriptsize Qwen3-4B} \\
\midrule
GRPO          & 22.7 & 32.3 & 34.6 & 52.7 \\
Entropy Bonus & 22.5 & 32.8 & 34.6 & 52.7 \\
EVOL-RL       & 22.7 & 33.8 & 39.2 & 53.0 \\
G$^2$RL       & 23.2 & 35.9 & 40.2 & 54.1 \\
\rowcolor{ourpurple}
\textbf{DiRL (Ours)} & \textbf{25.8} & \textbf{39.9} & \textbf{41.4} & \textbf{58.3} \\
\bottomrule
\end{tabular}%
}
\end{table}

\begin{figure*}[!t]
    \centering
    \includegraphics[scale=0.23]
    {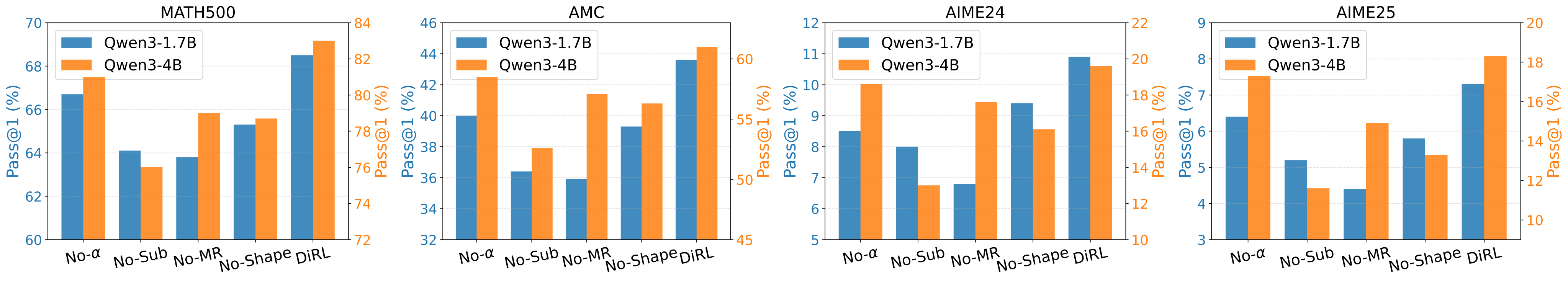}
    \caption{Ablation studies on four datasets.}
    \label{results:ablation}
\end{figure*}

\subsection{Main Results}
Table~\ref{tab:main_results} summarizes the performance of DiRL and baselines on four datasets for Qwen3 (1.7B and 4B) backbones. 
On both backbones, DiRL consistently outperforms baselines in all cases. 

For the Qwen3-1.7B, DiRL reaches $68.5$ pass@1 on MATH500 and $43.6$ on AMC, with similar gains extending to the harder AIME problems (pass@1 of $10.9$ on AIME24 and $7.3$ on AIME25) across maj@16 and pass@16. The consistent gains indicate that the direction-aware reward shaping progressively pulls the policy toward reasoning and away from memorization patterns. 
These trends amplify on Qwen3-4B. DiRL reaches $61.0$ pass@1 on AMC with parallel gains on maj@16 ($+9.3$) and pass@16 ($+8.1$), and achieves $83.0$ pass@1 on MATH500, $19.6$ on AIME24, and $18.3$ on AIME25. 
The gains suggest that the reasoning-anchored exploration allows the policy to cover a wider range of valid reasoning paths. 
Across both model sizes, DiRL improves pass@1 on every dataset, demonstrating that direction-aware exploration consistently moves probability mass toward higher-quality solutions.

\subsection{Generalization Results}
\label{sec:gen_results}

To evaluate the generalizability of DiRL beyond mathematical reasoning, we further conduct experiments on two broad-coverage reasoning benchmarks, GPQA \cite{rein2023gpqa} and MMLU-Pro \cite{wang2024mmlu}. 
The results are reported in Table~\ref{tab:generalization}, where DiRL consistently achieves the best performance across benchmarks and model scales.

For GPQA, DiRL improves pass@1 from 23.2 to 25.8 on Qwen3-1.7B and from 35.9 to 39.9 on Qwen3-4B, outperforming the strongest baseline G$^2$RL by 2.6 and 4.0 absolute points, respectively. Similar trends are observed on MMLU-Pro, where DiRL raises pass@1 from 40.2 to 41.4 on Qwen3-1.7B and from 54.1 to 58.3 on Qwen3-4B, yielding gains of 1.2 and 4.2 points over G$^2$RL.  
These results demonstrate that the benefits of DiRL are not confined to mathematical problem solving. By explicitly promoting reasoning-aligned exploration, DiRL learns policies that enhance the model's general reasoning ability, rather than overfitting to a specific benchmark or task format. 

\subsection{Ablation Study}
\label{sec:ablation_results}
We further conduct an ablation study to evaluate the contribution of each component in DiRL to the performance gain. We deactivate different components and form the following variants. 
\textbf{No-}$\bm{\alpha}$ replaces the weight $\alpha_t$ in Eq.(\ref{weighted-eq}) with uniform weights, 
\textbf{No-Sub} removes the $\mathcal{G}_R/\mathcal{G}_M$ partition and the exploration score $\nu^{(i)}$ uses the full group $\mathcal{G}$ as $\mathcal{R}_i$, 
\textbf{No-MR} makes the reference set $\mathcal{R}_i$ for memorization responses replaced by the $\mathcal{G}_M \setminus \{i\}$, 
and \textbf{No-Shape} removes the sign flipping in Eq.(\ref{eq12}).

The results are shown in Figure \ref{results:ablation}. 
All components are useful for DiRL as removing any one of them results in a performance decline.  
Among these, No-Sub exhibits substantial and consistent drops across datasets, 
indicating that explicitly separating reasoning and memorization is the foundation of DiRL. Without this partition, the exploration objective cannot distinguish beneficial reasoning trajectories from memorization-driven ones.
We also observe performance declines for No-MR, which highlights the importance of evaluating memorization-aligned responses relative to reasoning trajectories. 
This design provides a meaningful optimization target that encourages memorization responses to move toward more reasoning-aligned behaviors. 
Similarly, No-Shape brings pronounced performance drops, showing that merely identifying reasoning versus memorization is insufficient. The model needs to actively encourage reasoning exploration and suppress memorization to guide policy updates effectively. 
Furthermore, No-$\alpha$ shows lower performance, demonstrating that token-level weighting further refines the directional signal by emphasizing reasoning-relevant tokens. 
Overall, the full DiRL consistently achieves the best results, validating the effectiveness of our reasoning anchored exploration.

\subsection{Further Analysis}

\begin{figure}[!t]
    \centering
    \includegraphics[scale=0.21]
    {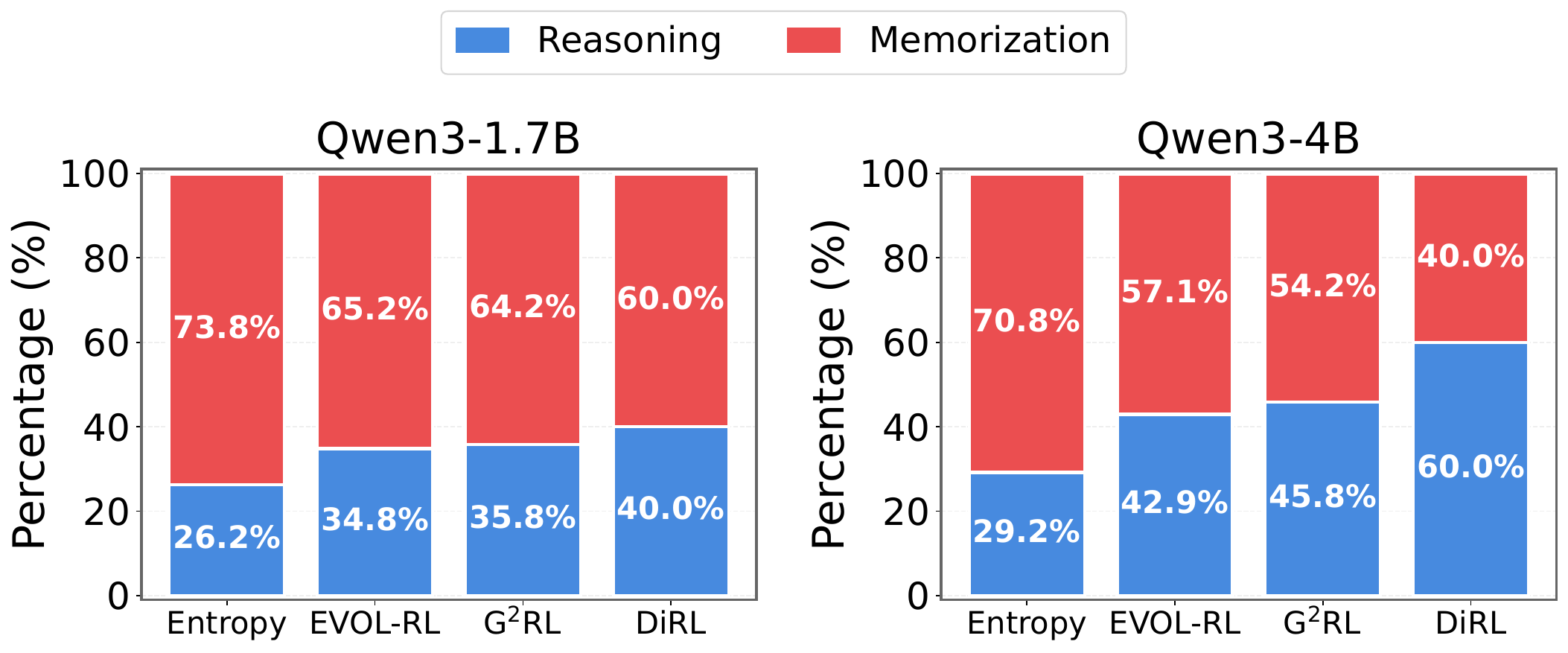}
    \caption{Reasoning- and Memorization-Aligned Response Ratios on AIME25.}
    \label{results:rm_ratio}
\end{figure}

\paragraph{Reasoning vs. Memorization Analysis.}
Since DiRL is explicitly designed to steer exploration toward reasoning, we further analyze the composition of model rollouts on AIME25. As shown in Figure~\ref{results:rm_ratio}, the proportion of reasoning-aligned responses increases consistently and reaches its highest level under DiRL across both model scales. Compared with G$^2$RL, DiRL raises the reasoning ratio from 35.8\% to 40.0\% on Qwen3-1.7B and from 45.8\% to 60.0\% on Qwen3-4B. 
And the higher reasoning ratio observed on Qwen3-4B suggests that larger models possess stronger latent reasoning capacity, allowing them to benefit more from reasoning-aligned exploration. 
These results demonstrate that DiRL pushes rollouts toward reasoning across model scales, validating the efficacy of the proposed direction-aware mechanism. 


\begin{table}[t]
\centering
\small
\setlength{\tabcolsep}{3pt}
\renewcommand{\arraystretch}{1.0}
\caption{Reasoning capability evaluation on GSM-Symbolic and its variants, reported in pass@1 (\%).}
\label{tab:gsm_symbolic}
\begin{tabular*}{\columnwidth}{@{\extracolsep{\fill}}llccc@{}}
\toprule
Model & Benchmark & EVOL-RL & G$^2$RL & DiRL \\
\midrule
\multirow{3}{*}{\shortstack{Qwen3\\-1.7B}}
& GSM-Sym.    & 67 & 67 & \textbf{77} \\
& GSM-Sym.-P1 & 69 & 69 & \textbf{71} \\
& GSM-Sym.-P2 & 34 & 46 & \textbf{47} \\
\midrule
\multirow{3}{*}{\shortstack{Qwen3\\-4B}}
& GSM-Sym.    & 85 & 80 & \textbf{93} \\
& GSM-Sym.-P1 & 81 & 80 & \textbf{89} \\
& GSM-Sym.-P2 & 78 & 76 & \textbf{82} \\
\bottomrule
\end{tabular*}
\end{table}

\paragraph{Reasoning Capability Analysis.}
To further assess whether DiRL improves genuine reasoning rather than fitting benchmark-specific patterns, we evaluate the trained models on GSM-Symbolic, GSM-Symbolic-P1 and -P2 \cite{mirzadeh2025gsm}, which introduce symbolic perturbations to GSM-style math problems. It provides a stricter test of whether the model can apply the underlying reasoning procedure under distributional variations. 
The results are reported in Table~\ref{tab:gsm_symbolic}. One can see that DiRL consistently achieves the best performance in all cases, showing that the gains observed on standard reasoning benchmarks remain robust under systematic symbolic perturbations. This indicates that the improvements of DiRL are not limited to the original evaluation distribution, but reflect a stronger ability to preserve the underlying reasoning process when the surface forms of problems are altered. 
The advantage of DiRL is particularly informative on P2, which contains stronger perturbations and therefore places greater emphasis on applying the underlying reasoning process. 
These results provide further evidence that the proposed direction-aware exploration strengthens the model's reasoning capability, rather than simply increasing task-specific accuracy.

\begin{figure}[!t]
    \centering
    \includegraphics[scale=0.35]
    {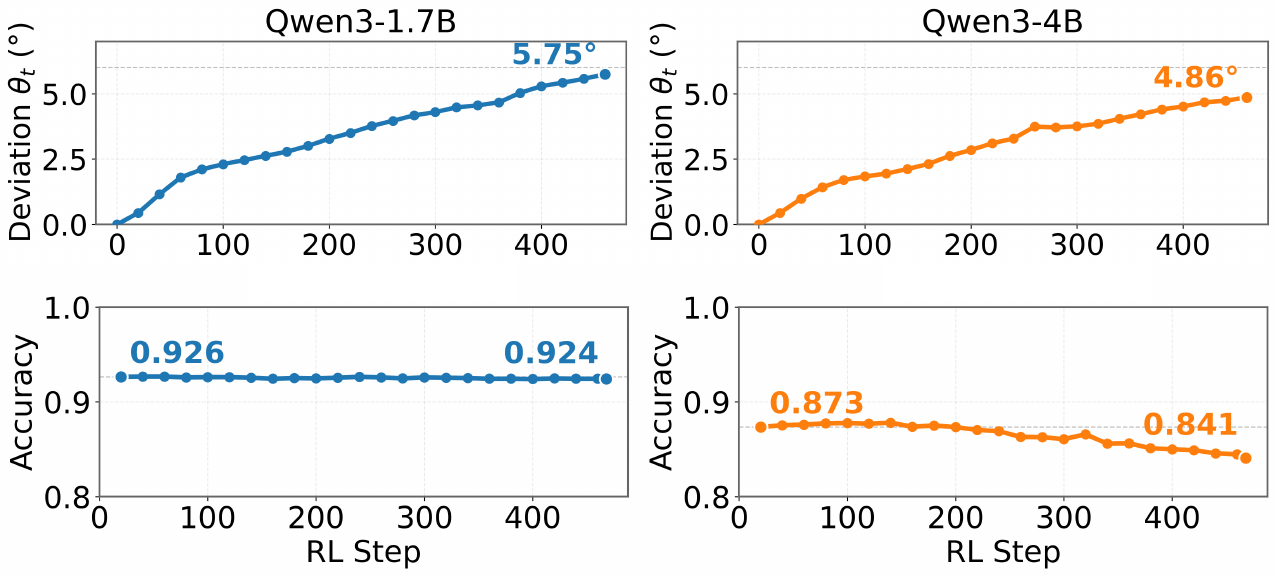}
    \caption{Geometric and semantic stability of the reasoning-memorization direction during RL training.}
    \label{results:dierction}
\end{figure}

\paragraph{Direction Analysis.}
A core design of DiRL is to use the reasoning-memorization direction as a fixed anchor throughout RL training. 
We evaluate its stability from both geometric and semantic perspectives. 
For geometric stability, we re-compute the direction $\mathbf{d}_t$ at each checkpoint and measure its angular deviation from the initial direction $\mathbf{d}_0$ as $\theta_t = \arccos \left( \mathbf{d}_t^\top \mathbf{d}_0 / (\|\mathbf{d}_t\| \, \|\mathbf{d}_0\|) \right)$. 
This metric isolates directional rotation from magnitude changes, where a smaller angle directly indicates higher geographical alignment \cite{park2023linear}. 
As shown in the upper panels of Figure~\ref{results:dierction}, the drift remains small throughout training, ending at only $5.75^\circ$ for Qwen3-1.7B and $4.86^\circ$ for Qwen3-4B. 
We further assess semantic stability by applying the fixed initial direction to hidden states from each checkpoint and testing whether it still distinguish reasoning examples from memorization examples. 
The lower panels show stable classification accuracy, changing from $0.926$ to $0.924$ for Qwen3-1.7B and remaining above $0.84$ for Qwen3-4B. 
These results show that the direction remains stable both geometrically and semantically as the policy evolves. 
It supports our choice to construct the direction only once before RL training, eliminating redundant computational overhead. 

\begin{figure}[!t]
    \centering
    \includegraphics[scale=0.57]
    {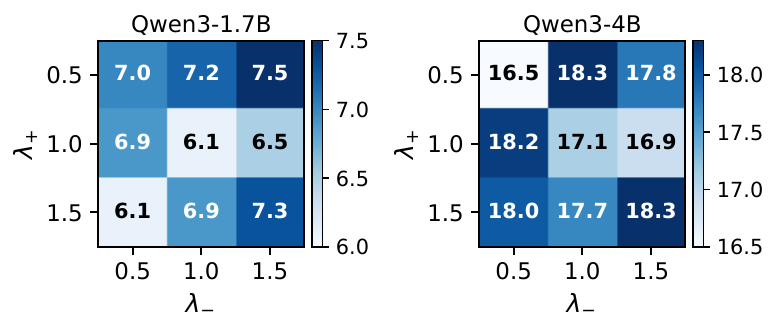}
    \caption{Effect of reward shaping hyperparameters on AIME25, reported in pass@1 (\%).} 
    \label{fig:hyper}
\end{figure}

\paragraph{Hyperparameter Analysis.}
We evaluate the sensitivity of DiRL to the reward shaping coefficients $\lambda_{+}$ and $\lambda_{-}$, which control the shaping strength for reasoning and memorization responses, respectively. We independently vary both coefficients over $\{0.5, 1.0, 1.5\}$ and report pass@1 on AIME25 for both Qwen3-1.7B and 4B. 
As shown in Figure~\ref{fig:hyper}, DiRL consistently achieves strong performance across all hyperparameter combinations, with small variation across different settings. 
It indicates that DiRL is not sensitive to the choice of coefficients. 
And we can see a trend that when $\lambda_{+}$ is fixed, increasing $\lambda_{-}$ often improves performance, suggesting that stronger shaping on memorization trajectories more effectively suppresses uninformative exploration. 
Similarly, increasing $\lambda_{+}$ under a fixed $\lambda_{-}$ tends to improve performance by providing a stronger signal for reasoning-aligned exploration. 
Overall, $(\lambda_{+}, \lambda_{-}) = (1.5, 1.5)$ delivers the strongest performance across model scales and is adopted as the default setting in all experiments.

\paragraph{Efficiency Analysis.}
In practice, the computational overhead introduced by DiRL over GRPO is remarkably lightweight. 
The auxiliary operations include computing token weights $\alpha_t$, partitioning responses, and calculating exploration scores, all of which are derived entirely from intermediate activations of the standard forward pass, bypassing the need for additional backward passes. In addition, the reasoning-memorization direction is pre-extracted before RL training and remains static throughout optimization. 
This efficiency is empirically validated in Table~\ref{tab:efficiency}, which reports the average training time per step for Qwen3-1.7B and 4B. One can see that DiRL induces only a modest training slowdown relative to GRPO, and this overhead exhibits stable scaling behavior as the model size increases. This underscores DiRL's capability to yield significant performance improvements with minimal additional computational cost.


\begin{table}[t]
\centering
\small
\setlength{\tabcolsep}{4pt}
\renewcommand{\arraystretch}{1.0}
\caption{Training efficiency of DiRL. Relative time is normalized by GRPO for each model scale.} 
\label{tab:efficiency}
\begin{tabular*}{\columnwidth}{@{\extracolsep{\fill}}llcc@{}}
\toprule
Model & Method & Sec./Step $\downarrow$ & Rel. Time $\downarrow$ \\
\midrule
\multirow{2}{*}{\shortstack{Qwen3\\-1.7B}}
& GRPO & 35.7 & 1.00$\times$ \\
& DiRL(Ours) & 40.2 & 1.13$\times$ \\
\midrule
\multirow{2}{*}{\shortstack{Qwen3\\-4B}}
& GRPO & 68.4 & 1.00$\times$ \\
& DiRL(Ours) & 80.5 & 1.18$\times$ \\
\bottomrule
\end{tabular*}
\end{table}

\section{Conclusion}
In this paper, we propose DiRL, a direction-aware reinforcement learning framework that anchors exploration to the internal reasoning-memorization distinction of LLMs. Unlike existing exploration methods that reward diversity indiscriminately in semantic or gradient spaces, DiRL selectively amplifies exploration aligned with reasoning while suppressing memorization-driven variations. 
Extensive experiments across mathematical and general reasoning benchmarks demonstrate that DiRL consistently improves reasoning performance, increases the proportion of reasoning-aligned rollouts, and generalizes better under symbolic perturbations. 
These findings suggest that effective exploration in LLM reinforcement learning should focus not merely on generating diverse trajectories, but on encouraging trajectories that genuinely expand reasoning. 
We hope this work motivates future research on representation-aware exploration, where the internal geometry of LLMs can be leveraged to guide reinforcement learning toward more robust reasoning behavior.

\section*{Limitations}

Like most reinforcement learning methods for large language models, DiRL is evaluated primarily on mathematical reasoning benchmarks and with a fixed reasoning-memorization direction extracted before training. Although the experiments demonstrate that this direction remains highly stable throughout optimization and consistently improves reasoning performance, its effectiveness may depend on the availability of representative data for constructing the initial direction. In addition, the current formulation focuses on a single global direction that captures the dominant contrast between reasoning and memorization. More complex tasks may involve multiple latent dimensions of reasoning behavior, such as planning, symbolic manipulation, and factual recall. Extending DiRL to automatically discover and exploit a richer set of reasoning-related directions is a promising direction for future work. 





\bibliography{custom}

\appendix

\section{Gradient Factorization Through Final-Layer Features}
\label{app:factorization}

We briefly explain why the aggregated feature
$\Phi(x,y)$ in Eq.~(7) provides a meaningful proxy
for comparing how different trajectories influence
parameter updates.

For an autoregressive language model, let
$\mathbf{h}_t \in \mathbb{R}^d$ denote the final-layer
hidden state at decoding step $t$.
The output logits are computed as
\begin{equation}
\mathbf{z}_t = W^\top \mathbf{h}_t,
\end{equation}
where $W \in \mathbb{R}^{V \times d}$ is the LM head.
Given the output distribution
$\mathbf{p}_t=\mathrm{softmax}(\mathbf{z}_t)$
and the realized token $y_t$,
the token-level log-likelihood gradient satisfies
\begin{equation}
\frac{\partial \log p_\theta(y_t \mid x,y_{<t})}
{\partial \mathbf{h}_t}
=
W\bigl(\mathbf{e}(y_t)-\mathbf{p}_t\bigr)
=
\boldsymbol{\phi}_t.
\end{equation}

Thus, $\boldsymbol{\phi}_t$ corresponds to the
first-order sensitivity of the token likelihood
with respect to perturbations in the final-layer
representation.
DiRL aggregates these token-level features using
the direction-aware weights $\alpha_t$:
\begin{equation}
\boldsymbol{\Phi}(x,y)
=
\sum_{t=1}^{L}
\alpha_t \boldsymbol{\phi}_t.
\end{equation}

For parameters $\theta_k$ at an arbitrary layer $k$,
the chain rule gives
\begin{equation}
\nabla_{\theta_k}\ell(x,y)
=
\mathcal{L}_k(x,y)\,
\boldsymbol{\Phi}(x,y),
\end{equation}
where $\mathcal{L}_k(x,y)$ is a trajectory-dependent
linear operator determined by the network Jacobian.
Therefore, parameter updates at every layer factor
through the aggregated feature $\boldsymbol{\Phi}(x,y)$.

Consequently, angular comparisons among
$\boldsymbol{\Phi}$ vectors provide a principled and
computationally efficient proxy for comparing how
different trajectories influence parameter updates.
Importantly, $\boldsymbol{\Phi}$ is computed entirely
from the forward pass without requiring additional
backpropagation.

\section{Training Hyperparameters}
\label{app:hyperparams}

Table~\ref{tab:hyperparams} lists all training hyperparameters used in our experiments. The same values are applied to all baselines and to DiRL. The maximum response length differs between backbones to accommodate the longer chain-of-thought outputs produced by the 4B model. And all RL training is conducted on 4 NVIDIA H100 (80GB) GPUs.

\begin{table}[h]
\centering
\caption{Training hyperparameters.} 
\small
\setlength{\tabcolsep}{14pt}
\renewcommand{\arraystretch}{1.25}
\begin{tabular}{@{}lr@{}}
\toprule
\textbf{Hyperparameter} & \textbf{Value} \\
\midrule
\multicolumn{2}{@{}l}{\textit{Optimization}} \\[2pt]
\quad Learning rate                & $5 \times 10^{-7}$ \\
\quad KL coefficient $\beta$       & $0.001$ \\
\quad Train batch size             & $16$ \\
\quad PPO mini-batch size          & $16$ \\
\quad PPO micro-batch size         & $2$ \\
\midrule
\multicolumn{2}{@{}l}{\textit{Rollout}} \\[2pt]
\quad Group size $m$               & $16$ \\
\quad Generation temperature       & $1.0$ \\
\quad Validation temperature       & $0.8$ \\
\quad Max response length (1.7B)   & $8{,}192$ \\
\quad Max response length (4B)     & $12{,}288$ \\
\midrule
\multicolumn{2}{@{}l}{\textit{DiRL-specific}} \\[2pt]
\quad Shaping strength $\lambda$   & $1.0$ \\
\quad Reward clip bound $c$        & $3$ \\
\bottomrule
\end{tabular}

\label{tab:hyperparams}
\end{table}

\section{Construction of the MATH Reasoning--Memory Split}
\label{app:math_split}

This appendix describes how we construct the MATH-R and MATH-M subsets used to extract the reasoning--memorization direction $\mathbf{k}$.
We treat reasoning and memorization as two ends of a continuous spectrum and use an LLM-as-a-judge protocol to assign each MATH problem a scalar score in $[0,1]$, where larger values indicate stronger reasoning requirements.

\paragraph{Labeling protocol.}
We apply GPT-4o to all 7,500 training problems in the MATH dataset.
For each problem, the model outputs a short justification and a reasoning score in $[0,1]$.
Since all MATH problems involve mathematical content, a generic reasoning--memory rubric may over-score most examples as reasoning-intensive.
We therefore use a MATH-adapted rubric that calibrates the two ends within mathematical problems: lower scores correspond to direct formula or template application, whereas higher scores correspond to problems requiring non-obvious constructions, multi-step deductions, case analysis, invariants, or other non-routine reasoning.
The full prompt is shown in Figure~\ref{fig:math_label_prompt}.

\paragraph{Partitioning rule.}
Let $q(x)\in[0,1]$ denote the GPT-4o reasoning score assigned to a MATH problem $x$.
We partition the dataset at threshold $0.5$:
$\text{MATH-R}=\{x:q(x)>0.5\}$, 
$\text{MATH-M}=\{x:q(x)\leq 0.5\}$. 
MATH-R is added to the reasoning-intensive set $\mathcal{D}_{+}$, while MATH-M is added to the memory-intensive set $\mathcal{D}_{-}$.

\paragraph{Use of labels.}
These GPT-4o labels are used only once, before RL training, to construct the contrastive sets for extracting the fixed direction $\mathbf{k}$.
They are not used as supervision for the policy, are not included in the verifier reward, and do not enter the GRPO objective.
During RL training, rewards are still determined by the task verifier; DiRL only uses the fixed direction $\mathbf{k}$ to compute direction-aware features and reward shaping terms.

\clearpage
\begin{figure*}[!t]
\centering
\begin{promptbox}{MATH Reasoning--Memory Labeling Prompt}

\textbf{System instruction:}
You are an expert math problem analyst.
Place the math problem below on a reasoning--memory spectrum.
Return:
\begin{itemize}[leftmargin=1.5em,itemsep=1pt,topsep=2pt]
    \item Concise justification (1--2 sentences).
    \item Score $[0.0, 1.0]$.
\end{itemize}

\textbf{Score definition:}
\begin{itemize}[leftmargin=1.5em,itemsep=1pt,topsep=2pt]
    \item \textbf{0.0 = Pure memorization:}
    a single, direct application of one standard formula, identity, or theorem, e.g., plugging into the quadratic formula, area formula, or distance formula.
    A student who has memorized the formula can answer in one step with no insight required.

    \item \textbf{1.0 = Pure reasoning:}
    requires non-obvious construction, such as an auxiliary line, auxiliary variable, or lemma; 4+ chained deductive steps; case analysis; invariant or extremal argument; clever substitution; or a non-standard re-framing of the problem.
    No single textbook formula yields the answer.

    \item \textbf{Intermediate values}
    reflect hybrid characteristics, such as mostly templated problems with one non-trivial step.
\end{itemize}

\textbf{Scoring guidelines} (additive, clamp to $[0.0, 1.0]$):
\begin{itemize}[leftmargin=1.5em,itemsep=1pt,topsep=2pt]
    \item Start from $0.5$ for a typical mid-difficulty MATH problem.
    \item $-0.3$ if the problem reduces to a single textbook formula or identity plug-in.
    \item $-0.2$ if the problem is a routine template, such as factoring, expanding, simplifying, basic arithmetic on given values, or direct substitution.
    \item $+0.15$ per non-trivial deductive step beyond the first, where each step requires choosing among multiple possible moves rather than simply executing the next line of an algorithm.
    \item $+0.2$ if the problem requires a non-obvious auxiliary construction, such as an extra variable, extra line, lemma, or change of basis.
    \item $+0.2$ if the problem requires case analysis or an invariant, extremal, pigeonhole, or parity argument.
    \item $+0.15$ if the problem requires recognizing and proving a pattern, such as induction, telescoping, or generating function.
    \item $-0.1$ if the answer is a single small integer obtainable by direct computation.
    \item $+0.1$ if the problem hides its structure, where the natural first attempt does not work and a re-framing is needed.
\end{itemize}

\textbf{Important:}
\begin{itemize}[leftmargin=1.5em,itemsep=1pt,topsep=2pt]
    \item Do \textbf{not} score based on the topic label, such as Algebra, Geometry, or Number Theory.
    A geometry problem can be 0.1 if it only requires plugging into an area formula, or 0.95 if it requires a clever auxiliary construction.
    \item Do \textbf{not} inflate the score just because numbers or LaTeX are present.
    Every MATH problem has numbers; that alone should not move the score.
    \item Use the full range.
    Avoid clustering at $0.7$--$0.9$.
    If a problem is a one-formula plug-in, score it below $0.3$.
\end{itemize}

\textbf{Target input:}
\texttt{Current Problem: \{question\}}

\textbf{Target output:}
\texttt{Analysis: <one or two sentences>} \\
\texttt{Score: <float in [0.0, 1.0]>}

\end{promptbox}
\caption{Prompt used for GPT-4o-based MATH reasoning--memory labeling.}
\label{fig:math_label_prompt}
\end{figure*}

\end{document}